\documentclass[10pt,conference,a4paper]{IEEEtran}
\ifCLASSINFOpdf
\else
\fi
\usepackage{graphicx}
\usepackage{caption}
\usepackage{url}
\usepackage{subfigure}
\usepackage{amsmath}
\usepackage[justification=centering]{caption}

\begin{document}
%
\title{Automatic Content-aware Non-Photorealistic Rendering of Images}

\author{\IEEEauthorblockN{Akshay Gadi Patil}
\IEEEauthorblockA{Electrical Engineering\\
Indian Institute of Technology\\
Gandhinagar, India-382355\\
Email: akshay.patil@iitgn.ac.in}
\and
\IEEEauthorblockN{Shanmuganathan Raman}
\IEEEauthorblockA{Electrical Engineering \& Computer Science and Engineering\\
Indian Institute of Technology\\
Gandhinagar, India-382355\\
Email: shanmuga@iitgn.ac.in}
}


%


\maketitle

\begin{abstract}
Non-photorealistic rendering techniques work on image features and often manipulate a set of characteristics such as edges and texture to achieve a desired depiction of the scene. Most computational photography methods decompose an image using edge preserving filters and work on the resulting base and detail layers independently to achieve desired visual effects. We propose a new approach for content-aware non-photorealistic rendering of images where we manipulate the visually salient and the non-salient regions separately. We propose a novel content-aware framework in order to render an image for applications such as detail exaggeration, artificial blurring and image abstraction. The processed regions of the image are blended seamlessly for all these applications. We demonstrate that content awareness of the proposed method leads to automatic generation of non-photorealistic rendering of the same image for the different applications mentioned above.
\end{abstract}


%
\IEEEpeerreviewmaketitle

\section{Introduction}
Non-photorealistic rendering of images have traditionally been done globally on the entire image grid. We would like to ask the question: can we achieve much better rendering of the given image if we deal with the image content in a content-aware manner? Content-aware processing has aided application such as re-targeting (\cite{mansfield2010scene},\cite{avidan2007seam}) and visual tracking \cite{yang2009context}. We would like to explore the possibility of  modern image filters with content-aware processing for more effective non-photorealistic rendering of images automatically. The challenge is to design a common pipeline for applications such as detail exaggeration, image abstraction and artificial blurring.\newline

Consider an image which needs to be manipulated by an artist in a content-aware manner. She may want to alter certain aspects corresponding to the foreground region in the image without altering the other contents. She would like to alter it using certain image editing software according to her requirements and then use the resulting image for display purposes. To achieve this, she would have to manually select that portion of the image in the software every time. Further, the region needs to be manipulated using specific tools manually which is quite time consuming for multiple images. Instead, if the process of altering the foreground region with the desired manipulation is made automatic, then such a problem can be addressed efficiently. The idea is to make image manipulation software to be content-aware with no human effort, thereby increasing the processing speed for large amount of images. \newline

Recent advances in computational photography applications focus on filtering algorithms for image processing. The need for preserving the edges during the smoothing operation in an image led to the development of edge preserving filters. One such filter is the bilateral filter. Well known filters such as box filter, Gaussian filter, and Laplacian filter perform convolution across the edges in an image.  However, edges across low texture variations, i.e, weaker edges, if smoothened will give a cartoon like appearance to the image which is known as image abstraction. Notable applications involving bilateral filter are high dynamic range compression \cite{durand2002fast}, flash/no-flash fusion \cite{petschnigg2004digital}, tone management \cite{bae2006two} and non-photorealistic relighting \cite{fattal2007multiscale}, to name a few.\newline 

In this paper, we present three new applications given below. 

\begin{enumerate}
	
	\item content-aware detail exaggeration using guided filtering, 
	\item detail exaggeration in salient region with defocused background, and
	\item independent abstraction of salient and non-salient regions in the image. \newline
	
\end{enumerate}

We use guided filter for edge-aware processing of the images in this work (\cite{bhat2010gradientshop},\cite{he2013guided}) The reason behind using guided filter instead of the bilateral filter for detail exaggeration is that the edges are relatively better preserved across strong contrast changes and illumination transitions without the introduction of any halos . The user has the freedom to alter the ``look" of the image. The image can be manipulated to give a non-photorealistic rendering by selective image abstraction using the bilateral filter. Instead of enhancing the detail in the entire photograph as mentioned in \cite{bae2006two}, \cite{fattal2007multiscale}, \cite{fattal2009edge1}, \cite{farbman2008edge}, \cite{paris2011local} and \cite{gastal2011domain}, we propose a new application of edge preserving filter aiming to exaggerate the detail only in the most salient region in the image and at the same time defocusing the non-salient region to give a more pronounced look of the salient region. In our case, we have used images which contain a salient foreground object and a background scene. We process only the foreground region and the background scene is left unaltered and vice-versa.\newline

The main contributions of our work are listed below.
\begin{enumerate}
 \item The proposed application is content-aware, i.e, this approach is well suited to manipulate the visibly significant regions in an image keeping all the other regions unaltered.
	
 \item It is a novel \textit{application} based framework based on edge preserving filtering, defocus blurring, recoloring and stylization techniques that processes the brightness, contrast and texture information in a content-aware manner.
	
 \item Since the method does not make use of any scale space pyramids and involves processing in the same scale, it is computationally less expensive.\\
	
\end{enumerate}

We discuss the existing works, motivation leading to detail enhancement, defocusing approaches and image abstraction in section II. In section III, we describe the framework of our approach. We then present our results in section IV as a new application of edge preserving filters and image stylization, and rendering the manipulated image back in the original photograph using the state-of-the-art image compositing technique. We end the paper with conclusions and scope for future work in section V.\newline

\section{Related Work}

 We derive our motivation from a rich body of existing works on edge preserving filters and their applications to images such as detail enhancement (\cite{fattal2009edge1}, \cite{fattal2007multiscale}, \cite{bae2006two}, \cite{farbman2008edge}, \cite{paris2011local} and \cite{gastal2011domain}), defocus blur \cite{subbarao1994depth}, and image abstraction (\cite{winnemoller2006real} and \cite{winnemoller2011xdog}), all using different signal processing tools which are explained below.\\ 
 
Bae \textit{et al.} in \cite{bae2006two} described a method for spatial detail variation. The amount of high frequency detail (texture) and its spatial variation is manipulated using a new \textit{textureness} map that performs an edge-preserving analysis.  Fattal \textit{et al.} in \cite{fattal2007multiscale} showed detail enhancement in images photographed with a fixed viewpoint but in different lighting conditions. They performed multi-scale decomposition of the images, applied the bilateral filter on them and combined the shading information across all the input images. A new method for edge-preserving multi-scale decomposition was proposed by Farbman \textit{et al.} in \cite{farbman2008edge}. They incorporate weighted least squares (WLS) optimization framework instead of the base-detail decomposition technique based on bilateral filter which supposedly are limited in their ability to extract details at arbitrary scales. Fattal demonstrated edge-preserving smoothing and detail enhancement using a new edge avoiding wavelet basis as explained in \cite{fattal2009edge1}. A new scheme for edge based image coarsening is proposed in \cite{fattal2009edge}. Here they construct a dimensionally reduced image space in which pixels are bound together according to the edge contents of the image using bilateral filter kernels. Gastal and Oliviera in \cite{gastal2011domain} propose a transform for edge preserving filter based applications. Bhat \textit{et al.} in \cite{bhat2010gradientshop} proposed a gradient domain optimisation framework for image and video processing which manipulates pixel differences (such as the first order image gradients) in addition to the pixel values of an image. 
 
 Blurring in an image can caused due to many reasons. Lens abberations, diffraction, turbulence, camera shake, defocus and fast moving object are some of the causes. Defocus blurring operation is an image smoothing operation. When we capture a scene using a camera, focusing is achieved by adjusting the focal length of a camera. Once a scene is captured, the amount of defocus or blur can be controlled, irrespective of the camera parameter settings, by making use of convolution operation in the spatial domain as presented in \cite{subbarao1994depth}. The spatial domain approach involves convolution of the image with a fixed or a spatially varying kernel. The most commonly used blurring kernel filter is the Gaussian filter. Defocus has been used in applications involving depth estimation \cite{subbarao1994depth}, video and image matting (\cite{mcguire2005defocus},\cite{mcguire2005defocus1}) and geometric shape estimation \cite{favaro2005geometric} with considerable success.\newline

  Image abstraction is the process of abstracting an image by suppressing the weaker edges while preserving the stronger ones iteratively. Decarlo and Santella developed a method to distinguish important parts in an image by drawing bold lines \cite{decarlo2002stylization}. But their approach needs user intervention and is computationally expensive for long video sequences containing many frames. Winnemoller in \textit{et al.} proposed a real time video and image abstraction method that modified the contrast in the luminance and the color features in the image.  Winnemoller proposed a new approach for stylistic depiction applications using the extended difference of Gaussians \cite{winnemoller2011xdog}. Image quantization after appropriately filtering the image would produce a good abstraction of the image since now the level of variations are fixed and any edges present after the filtering operation can take one of the quantized values \cite{winnemoller2006real}.
 
 Humans are smart to figure out the edges, illumination variations and the flat regions in an image. Over the years our visual attention system has evolved to in its ability to predict the most relevant features of a scene where our eyes fixate in a fixed-time, free-viewing scenario \cite{harel2006graph}. Itti \textit{et al.} in \cite{itti1998model} proposed a model of saliency based visual attention that results in a saliency map which is robust to noise. Harel \textit{et al.} in \cite{harel2006graph} proposed a new approach of visual attention fixation in an image based on Markov chain approach on graphs obtained by connecting pixels in an image and working on the similarity measure among the edges in the graph. The main aim of any saliency algorithm is to highlight the significant locations in an image that is informative according to some criterion, like the human fixation. Li \textit{et al.} in \cite{li2014secrets} addresses the design bias problems of the existing saliency algorithms that create discomforting disconnections between fixations and salient object segmentation. In other words, a saliency map may include areas that do not constitute the salient object, yet we use such algorithms because they give us a measure of the content-awareness in a given image.  

\section{Proposed Methodology}

\begin{figure*}[t]
	\includegraphics[height = 7.2 cm, width = 12.8 cm]{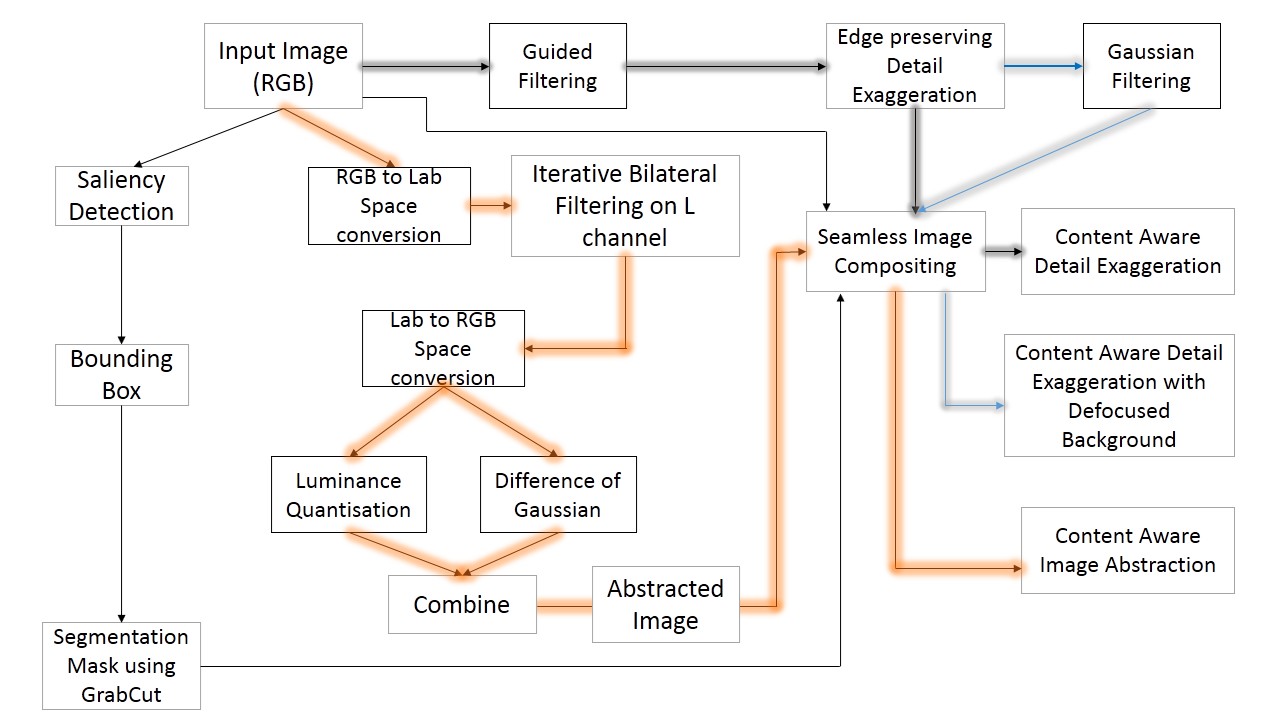}
	\centering
	\caption{Proposed Approach. \textit{The black line gives the flow for content-aware detail exaggeration, the blue line shows the flow for detail exaggeration with defocused background and the brown line gives the flow for content-aware image abstraction.} }
	\label{Figure 1}
	
\end{figure*}

 The proposed approach involves the processing of salient region in an image which is directed towards three applications in this paper. The content-aware processing for non-photorealistic image rendering for these applications is a novel contribution. A similar technology is used in the \textit{Smart Looks} plug-in in the Adobe Photoshop Elements 14 in a non context-aware manner\cite{adobe}. We are not aware of the technology behind this plug-in and any other significant work in this direction to the best of our knowledge. The diagram in Fig. \ref{Figure 1} explains our methodology of content-aware non-photorealistic rendering of images for the three applications mentioned in this paper.\newline
 
\begin{figure*}[htbp]
	
	\centering
	\begin{subfigure}[]
		\centering
		\includegraphics[width = 0.95 in,height = 0.71428 in]{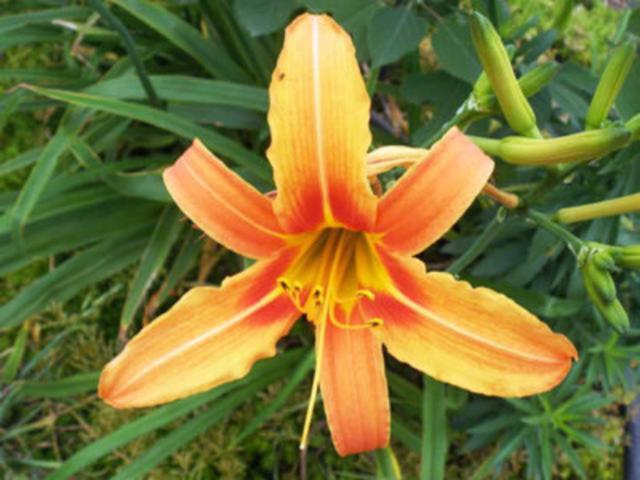}
	\end{subfigure}
	\begin{subfigure}[]
		\centering
		\includegraphics[width = 0.95 in,height = 0.71428 in]{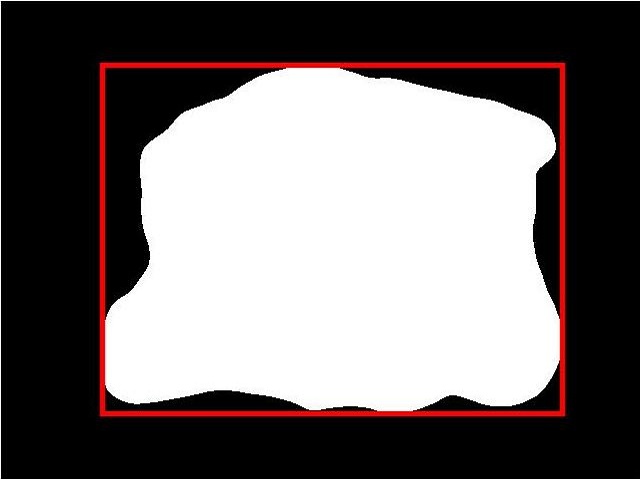}
	\end{subfigure}%
	\begin{subfigure}[]
		\centering
		\includegraphics[width = 0.95 in,height = 0.71428 in]{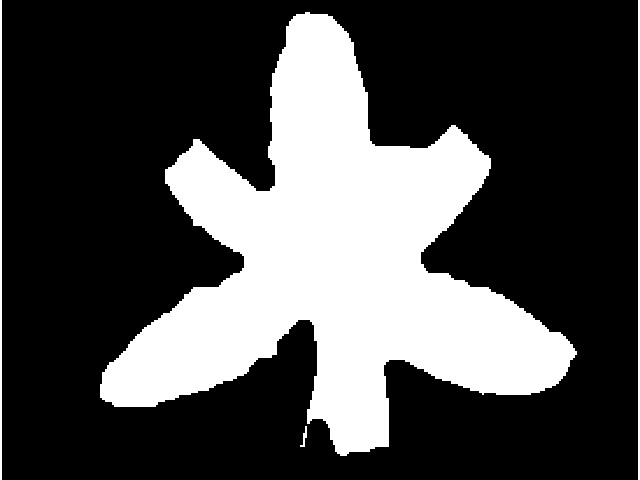}
	\end{subfigure}%
	\begin{subfigure}[]
		\centering
		\includegraphics[width = 0.95 in,height = 0.71428 in]{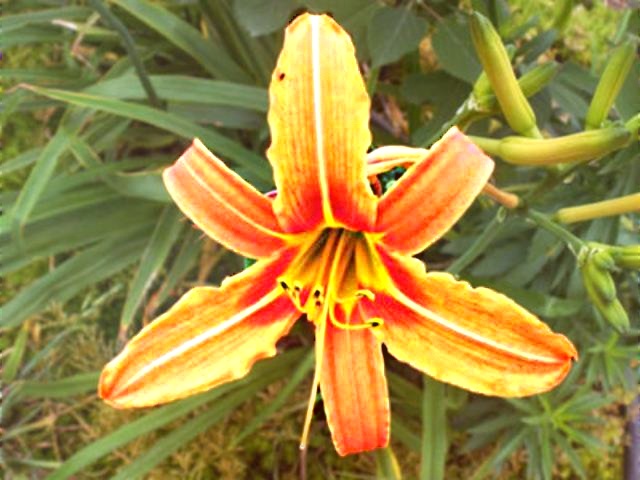}
	\end{subfigure}%
	\begin{subfigure}[]
		\centering
		\includegraphics[width = 0.95 in,height = 0.71428 in]{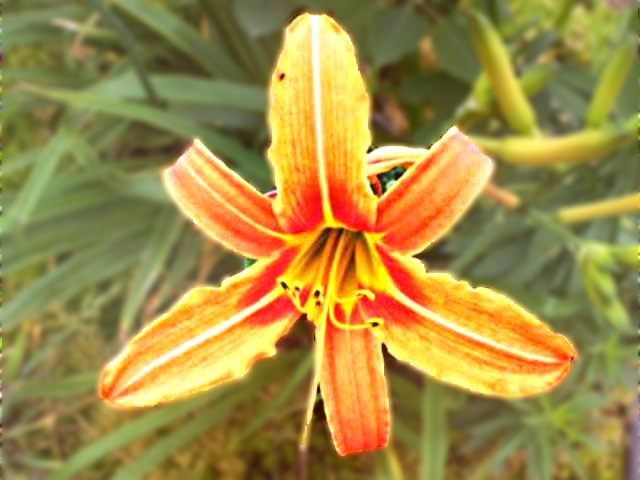}
	\end{subfigure}%
	\begin{subfigure}[]
		\centering
		\includegraphics[width = 0.95 in,height = 0.71428 in]{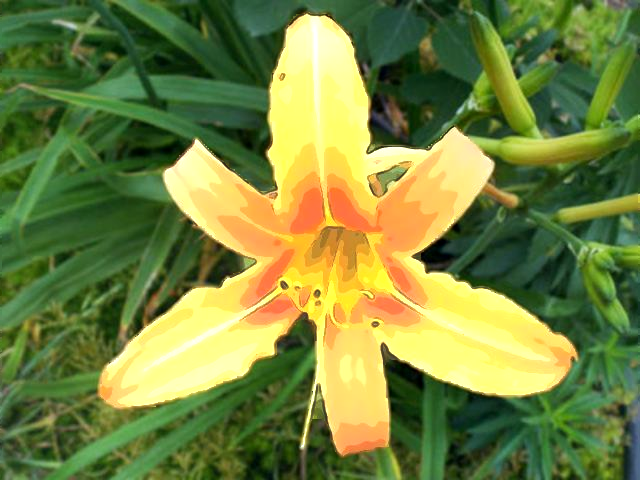}
	\end{subfigure}
	\begin{subfigure}[]
		\centering
		\includegraphics[width = 0.95 in,height = 0.71428 in]{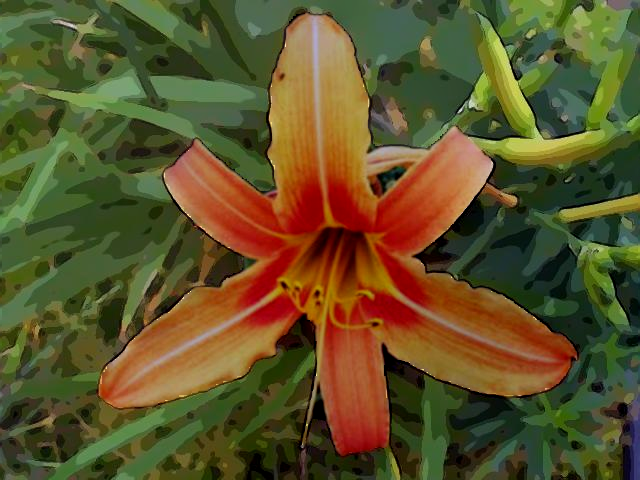}
	\end{subfigure}%
	\caption{ \textit{(a) Input image, (b) Bounding box around saliency mask, (c) Mask after GrabCut, (d) Content aware Detail exaggeration , (e) Content aware detail exaggeration with background defocussed, (f) Foreground region abstraction, (g) Background scene abstraction.}}
	\label{Figure2}
	
\end{figure*}

\subsection{Saliency Based Segmentation}

 As the title of our paper suggests, we aim to make our system content-aware. For this, we first find the salient region in the given image. We employed a graph based visual saliency method proposed by Harel \textit{et al.} in \cite{harel2006graph}. This helps us in narrowing down to identify the most visually salient region in an image. The images used in our approach were collected from the salient object dataset provided by Li \textit{et al.} in \cite{li2014secrets}. We obtain a binary mask from the saliency map using Otsu's threshold \cite{otsu1975threshold} as shown in Fig. \ref{Figure2}(b). However, the saliency mask may include portions of the image on which a human eye fixates but it may not be a part of the most salient region in the image. We find a bounding box around the binary mask corresponding to the salient region for each image automatically which is shown in red in Fig. \ref{Figure2}(b).   To ensure that the computer emulates the human perception to understand the visually meaningful parts in an image we make use of GrabCut technique to accurately extract the salient region without any background contribution for further processing \cite{rother2004grabcut}. \\

\subsection{Content-aware Processing and Compositing}
 The bounding box along with the input image of Fig. \ref{Figure2}(a) is given to the GrabCut algorithm proposed by Rother \textit{et al.} in \cite{rother2004grabcut} which extracts the complete salient region/object inside the bounding box. A mask of the object extracted using the GrabCut algorithm is shown in Fig. \ref{Figure2}(c). The input to the GrabCut algorithm should be an image containing the foreground object within the bounding box irrespective of whether or not the object of interest needs to be processed or manipulated with respect to its contents.\newline
 
 \subsubsection{Content-aware Detail Exaggeration}
 We employ an edge preserving filtering approach taking cues from the existing literature as explained in the related work section. The edge preserving filter is used to get a structure and texture decomposition of the image without any halos. The input image is filtered using a guided filter as explained in \cite{he2013guided}. Guided filter was proposed by He \textit{et al.} in \cite{he2013guided} where the output of the filtering operation is a local linear transform of the guidance image. The input image is enhanced with respect to the details present in it as can be seen from Fig. \ref{Figure2}(d). It can be observed that the image has a similar feel as the original image with the details exaggerated. Fundamentally, detail exaggeration requires one to manipulate the gradients in the detail layer (which is the difference between the input image and the base layer) obtained from the guided filter. The modified gradients need to be re-integrated into the image to (a) prevent gradient reversal, (b) for the manipulations to take place, and (c) for the effects to be visible. The use of bilateral filter introduces halos along edges with strong illumination changes on either side of it because of gradient reversal in the reconstruction step. The detail exaggerated image is obtained by combining the boosted detail layer with the base layer. Unlike bilateral filter, the guided filter does not suffer from gradient reversal artifacts \cite{he2013guided}. The input image of Fig. \ref{Figure2}(a) along with the saliency mask in Fig. \ref{Figure2}(c) and the detail exaggerated image is given as input to the image compositing algorithm proposed in \cite{tao2013error} to get the content-aware detail exaggerated image as shown in Fig. \ref{Figure2}(d).\newline
 
 \subsubsection{Content-aware Detail Exaggeration With Defocused Background }
 As we mentioned in the introduction section, we defocus the background and exaggerate only the salient region to give a more pronounced look of the enhanced image as could be seen from Fig. \ref{Figure2}(e). As mentioned before, there are spatially variant and spatially invariant blurring kernels for defocus operation. We use a simple approach to defocus the image using a fixed size Gaussian kernel. Every pixel $(x,y)$ in the image is operated upon with a Gaussian filter $G_{\sigma} (x,y)$ of kernel $9\times 9$ with a standard deviation 4 around the neighborhood of the pixel. The larger the value of $\sigma$, the larger is the blurring effect in the image. This filtering operation is given by Eq. \ref{eq 1} below:
 
 \begin{equation} \label{eq 1}
 \begin{aligned}
 \widehat I(x,y) =  I(x,y)  \otimes {G_{\sigma} (x,y)}
 \end{aligned}
 \end{equation}
 
 So the entire image is defocused. But we want only the background defocused. To achieve this, the defocus blurred image along with the mask in Fig. \ref{Figure2}(c) and the image in Fig. \ref{Figure2}(d) are given as input to the image compositing algorithm for a seamless compositing. We used the error tolerant image compositing algorithm proposed by Tao \textit{et al.} in \cite{tao2013error}. The output of this system is the content-aware detail exaggerated image with defocused background which is shown in Fig. \ref{Figure2}(e).\\
 
 \subsubsection{Content-aware Image Abstraction}
 Our next application is to show image abstraction in the salient and non-salient regions separately. We make use of the approach proposed in \cite{winnemoller2006real} for image abstraction. The input color image after conversion to $Lab$ space is filtered using the bilateral filter. For the Gaussian filter, we used a spatial kernel with standard deviation 3 and a range kernel with a standard deviation 0.1. It is converted back to RGB space. Luminance values in the resulting image are quantized into 10 different levels and difference-of-Gaussian filter is applied on the resulting RGB image. These two images are then combined to get the abstracted image which gives a cartoon like appearance to the image as can be seen from Fig. \ref{Figure2}(f) (foreground abstracted) and Fig. \ref{Figure2}(g)(background abstracted). To get the abstraction of the visually important region, we again employ the GrabCut technique combined with error tolerant image compositing algorithm. The input image in its entirety is abstracted using the method described above. Mask obtained after the GrabCut technique, along with the input image and the abstracted image is given as the input to image compositing algorithm proposed in \cite{tao2013error}. The result is that there is content-aware abstraction which can be seen from Fig. \ref{Figure2}(f). If the binary mask obtained after the GrabCut algorithm is inverted, and the above mentioned operations are performed with this new mask, then the non salient region in the image is abstracted as could be seen from Fig. \ref{Figure2}(g).\newline
 
\section{Results and Discussions}
\begin{figure*}[h]
    \centering
    \begin{subfigure}
    	\centering
    	\includegraphics[width = 1.6625 in,height = 1.25 in]{40}
    \end{subfigure}
    \begin{subfigure}
    	\centering
    	\includegraphics[width = 1.6625 in,height = 1.25 in]{40_defocus_and_exaggerated}
    \end{subfigure}%
    \begin{subfigure}
    	\centering
    	\includegraphics[width = 1.6625 in,height = 1.25 in]{40_salient_abstraction}
    \end{subfigure}
    \begin{subfigure}
    	\centering
    	\includegraphics[width = 1.6625 in,height = 1.25 in]{40_nonsalient_abstraction}
    \end{subfigure}%
    \\\vspace{0.02in}
	
	\centering
	\begin{subfigure}
		\centering
		\includegraphics[width = 1.6625 in,height = 1.2472 in]{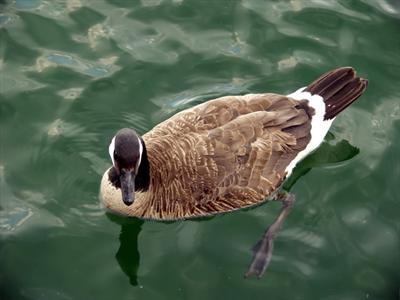}
	\end{subfigure}
	\begin{subfigure}
		\centering
		\includegraphics[width = 1.6625 in,height = 1.2472 in]{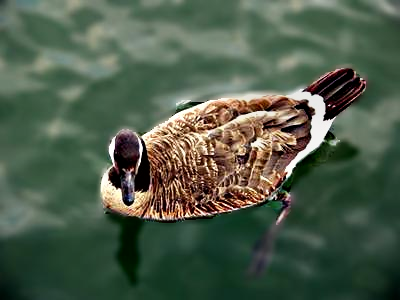}
	\end{subfigure}%
	\begin{subfigure}
		\centering
		\includegraphics[width = 1.6625 in,height = 1.2472 in]{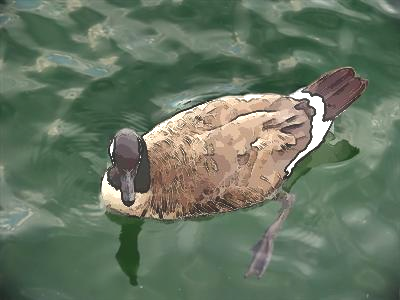}
	\end{subfigure}
	\begin{subfigure}
		\centering
		\includegraphics[width = 1.6625 in,height = 1.2472 in]{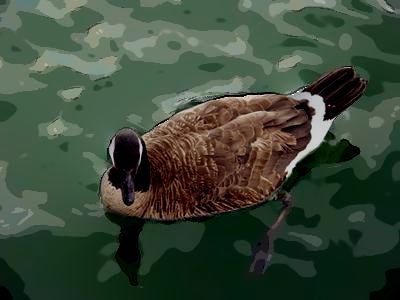}
	\end{subfigure}%
	 \\\vspace{0.02in}

\centering
\begin{subfigure}
	\centering
	\includegraphics[width = 1.6625 in,height = 1.1098 in]{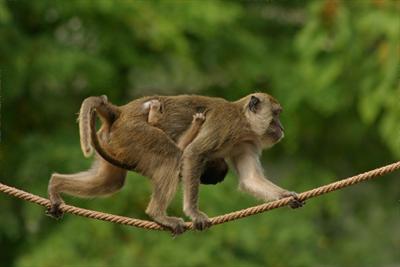}
\end{subfigure}
\begin{subfigure}
	\centering
	\includegraphics[width = 1.6625 in,height = 1.1098 in]{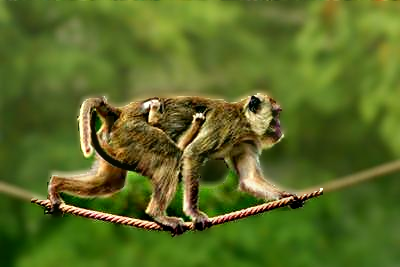}
\end{subfigure}%
\begin{subfigure}
	\centering
	\includegraphics[width = 1.6625 in,height = 1.1098 in]{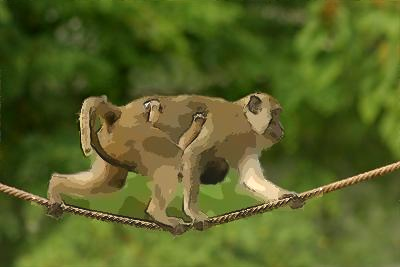}
\end{subfigure}
\begin{subfigure}
	\centering
	\includegraphics[width = 1.6625 in,height = 1.1098 in]{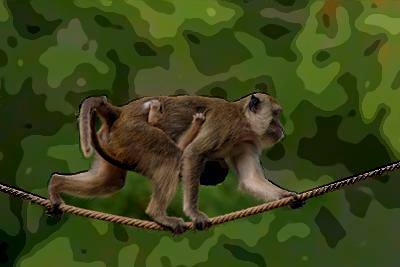}
\end{subfigure}%
    \\\vspace{0.02in}

    \centering
    \begin{subfigure}
    	\centering
    	\includegraphics[width = 1.6625 in,height = 1.25 in]{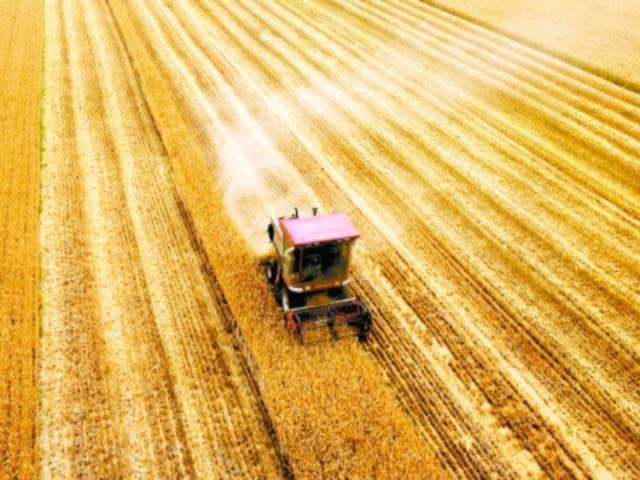}
    \end{subfigure}
    \begin{subfigure}
    	\centering
    	\includegraphics[width = 1.6625 in,height = 1.25 in]{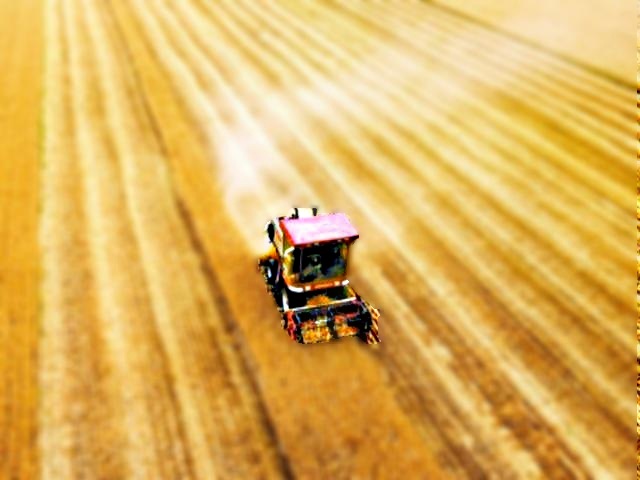}
    \end{subfigure}%
    \begin{subfigure}
    	\centering
    	\includegraphics[width = 1.6625 in,height = 1.25 in]{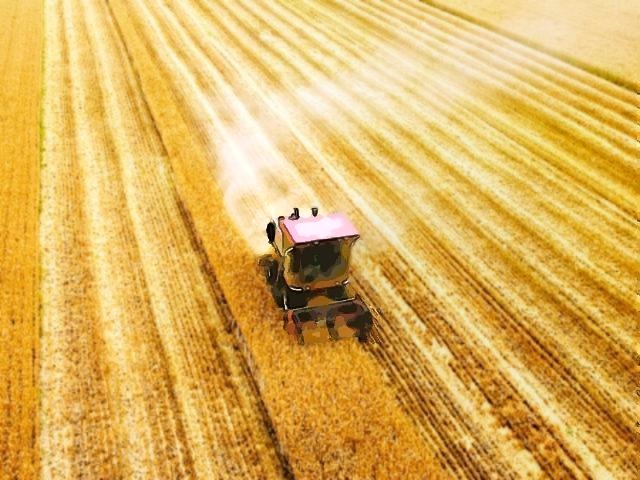}
    \end{subfigure}
    \begin{subfigure}
    	\centering
    	\includegraphics[width = 1.6625 in,height = 1.25 in]{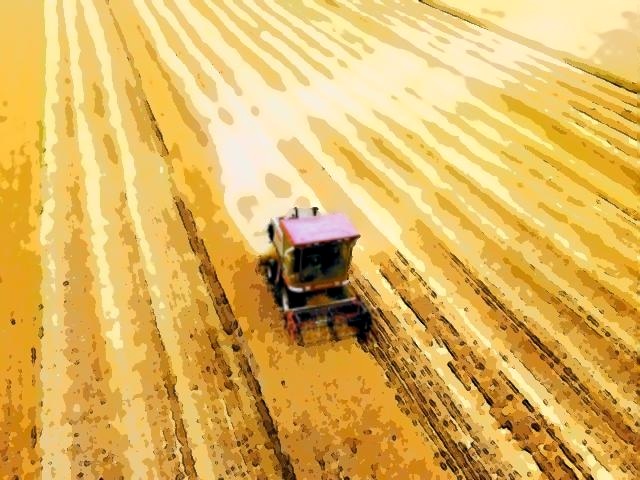}
    \end{subfigure}%
    \\\vspace{0.02in}

	\centering
	\begin{subfigure}
		\centering
		\includegraphics[width = 1.6625 in,height = 1.25 in]{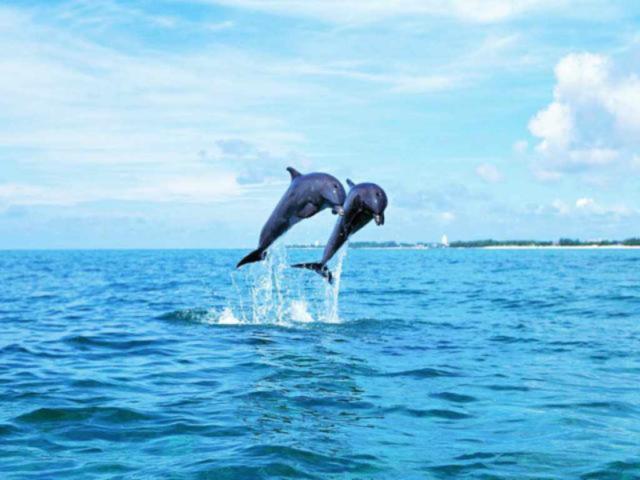}
	\end{subfigure}
	\begin{subfigure}
		\centering
		\includegraphics[width = 1.6625 in,height = 1.25 in]{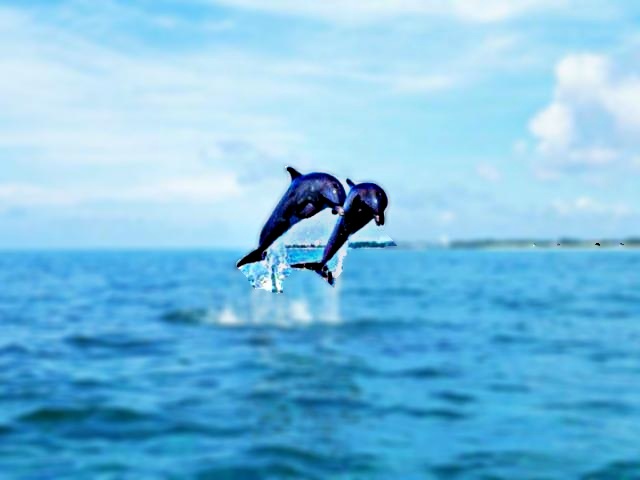}
	\end{subfigure}%
	\begin{subfigure}
		\centering
		\includegraphics[width = 1.6625 in,height = 1.25 in]{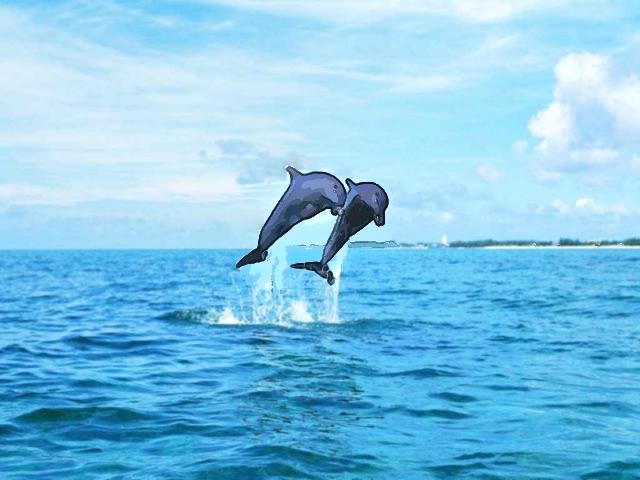}
	\end{subfigure}
	\begin{subfigure}
		\centering
		\includegraphics[width = 1.6625 in,height = 1.25 in]{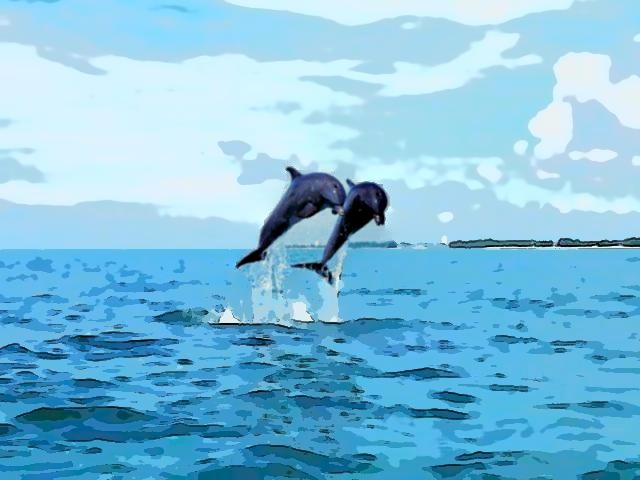}
	\end{subfigure}%
	\\\vspace{0.02in}

	\centering
	\begin{subfigure}[]
		\centering
		\includegraphics[width = 1.6625 in,height = 1.25 in]{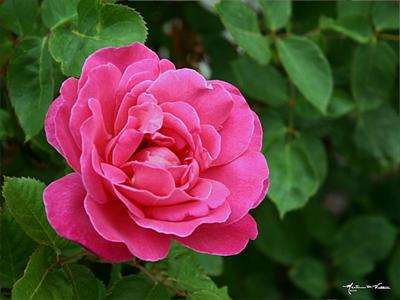}
	\end{subfigure}
	\begin{subfigure}[]
		\centering
		\includegraphics[width = 1.6625 in,height = 1.25 in]{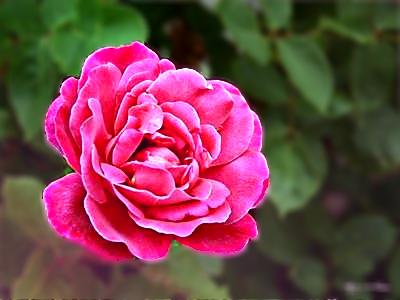}
	\end{subfigure}%
	\begin{subfigure}[]
		\centering
		\includegraphics[width = 1.6625 in,height = 1.25 in]{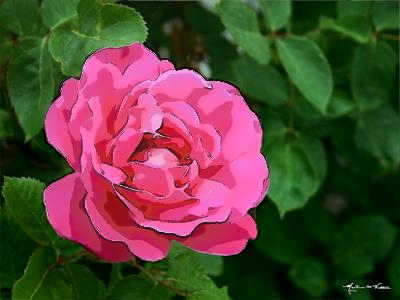}
	\end{subfigure}
	\begin{subfigure}[]
		\centering
		\includegraphics[width = 1.6625 in,height = 1.25 in]{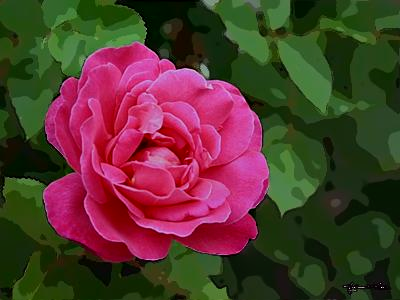}
	\end{subfigure}%

\caption{ \textit{Results:(a) Input image, (b) Detail Exaggeration with defocused background, (c) Foreground region abstraction, (d) Background scene abstraction.}}
\label{Figure3}

\end{figure*}

We present the results for a set of six images on the three different filtering applications mentioned in the application pipeline described in the previous section, which are presented from top row to the bottom row of Fig. \ref{Figure3}. For every image, we show the content-aware processed images along with the original image. As could be seen from Fig. \ref{Figure3}(b), the images exaggerated using the guided filter approach \cite{paris2006fast} produce good exaggeration and this approach was selected from a set of other detail exaggeration methods based on minimal artifacts in the processed image. The background is defocused to give a more pronounced look of the detail exaggerated in the salient region.  At the same time, the overall background and foreground illumination in the image is increased and has more contrast as seen in Fig. \ref{Figure3}(b) column. Column (c) of Fig. \ref{Figure3} shows images where the salient region is abstracted and the non salient region is left unaltered. There is illumination and contrast change happening only in the foreground region when the salient region is abstracted. The level of abstraction can be controlled by applying the bilateral filter iteratively to suit the requirements from the user end. The last column of Fig. \ref{Figure3}, i.e, Fig. \ref{Figure3}(d) shows the results of abstraction on the background scene. The illumination in the foreground region as compared to the original image is reduced. The robustness of the error tolerant image compositing technique \cite{tao2013error} ensures the seamless image compositing operation after the processing of the respective content-aware region. It also ensures that there is no background clutter effect and the halos do not appear during processing involved. Our approach is content-aware as it processes only the salient object in the image keeping the rest of the original image as it is. \newline
 
We performed the experiments in MATLAB environment on a laptop that runs Windows 8 with Intel core i5 (1.7 GHz) processor with 6 GB RAM. Typical time required for an image of size 800 $\times$ 533 for content-aware detail exaggeration is 30 seconds, for content-aware detail exaggeration with defocused background is 37 seconds, for content-aware image abstraction for both salient and non-salinet regions is 75 seconds each. Since we have designed our system to be content-aware, the kind of images our method is well suited for are the ones which contain a salient foreground region and a background scene.\newline

\section{Conclusions}
The proposed approach manipulates the details and processes the image in a content-aware manner, i.e, only the most salient object in the image is processed using edge preserving filtering. We do not decompose the input image into scale space pyramids for any of the addressed applications. An image compositing technique is used which takes the mask corresponding to foreground object in an image which has to be composited on a background scene. The proposed approach does not introduce any artifacts in the process of making the system content-aware, be it content-aware detail exaggeration with defocused background  or non salient image abstraction which inherently gives cartoon effect to the image. Such an application can be used for non-photorealistic rendering and can be extended to more applications requiring content awareness for various image manipulations. \newline

Future scope involves developing content-aware applications for other computational photography problems such as high dynamic range imaging and flash/no-flash photography. We aim to carry out qualitative analysis and image quality assessment of the results produced using the proposed approach. Subjective studies could also be carried out for determining the visual appeal of the abstracted images and thereby controlling the amount of abstraction suitable for non-photorealistic rendering. We also plan to explore other notions of content-aware processing other than saliency for the proposed applications. We  believe that the framework proposed will stimulate research on applications of automatic content-aware non-photorealistic rendering of images. \newline
  \addtocounter{subfigure}{-20}






%



\end{document}